# Towards a Semantic-based Approach for Modeling Regulatory Documents in Building Industry

K.R. Bouzidi & B. Fies
*Centre Scientifique et Technique du Bâtiment, Sophia Antipolis, France*

C. Faron-Zucker & N. Le Than
*I3S, Université Nice Sophia Antipolis and CNRS, France*

O. Corby
*INRIA Sophia Antipolis, France*

ABSTRACT: Regulations in the Building Industry are becoming increasingly complex and involve more than one technical area. They cover products, components and project implementation. They also play an important role to ensure the quality of a building, and to minimize its environmental impact. In this paper, we are particularly interested in the modeling of the regulatory constraints derived from the Technical Guides issued by CSTB and used to validate Technical Assessments. We first describe our approach for modeling regulatory constraints in the SBVR language, and formalizing them in the SPARQL language. Second, we describe how we model the processes of compliance checking described in the CSTB Technical Guides. Third, we show how we implement these processes to assist industrials in drafting Technical Documents in order to acquire a Technical Assessment; a compliance report is automatically generated to explain the compliance or noncompliance of this Technical Documents.

## 1 INTRODUCTION

Regulations in the Building Industry are becoming increasingly complex and involve more than one area. They cover products, components and project implementation. They also play an important role to ensure the quality of a building, its features, and to minimize its environmental impact. For 30 years, CSTB proves its expertise in this field through the development of the complete encyclopaedia of French technical and regulatory texts in the domain of the construction: the REEF. In the framework of a collaboration between CSTB and the I3S laboratory, we work on the acquisition of knowledge from technical and regulatory information contained in the REEF and the automated processing of this knowledge with the final goal of assisting professionals in the exploitation of these texts and the creation of new texts. We implement our work in CSTB to help in the writing of Technical Assessments. It is a question to specify how these assessments are created and how to standardize their structure using models and adaptive semantic services.

A Technical Assessment (in French: Avis Technique or ATec) is a document containing technical information on the usability of a product, material, component or element of construction, which has an innovative character. We chose Technical Assessments as a study model because CSTB has the mastership and a wide experience in these kinds of technical documents.

A Technical Assessment is drafted at the request of an industrial. The industrial starts by sending a request for technical assessment to relevant departments within the CSTB. Then the CSTB instructors transmit to the industrial the resolution of technical assessment and a form for developing the technical document. It is a preformatted Word file containing chapters, text and instructions on how to fill it out. When done, the document is supposed to describe with the right accuracy the process, product or material candidate for a Technical Assessment. This document is therefore studied by a specialized group who will be responsible for delivering the technical assessment. The industrial is supposed to fill in the form and send a complete document to CSTB.

We are particularly interested in this paper to model regulatory constraints that validates the Technical Assessment. In section 2 we describe the practical guides. In section 3 we present our approach to model regulatory constraints. In section 4 we detail our approach to interpret compliance checks results. Related work is discussed in section 5. We conclude in Section 6.

## 2 TECHNICAL GUIDES

Technical guides are regulatory complements offered by CSTB to industrials. They do not replace the texts, regulations (decree, circular), normative (DTU) or code (Atec). They enable an easier reading of technical rules and collect details of execution presenting a wide range of possible implementations. The main actors concerned by these guides are project owners, contractors, firms. The main purpose of these documents is to address issues of health and safety faced by industry in particular during the design phase.

There are many case studies, diagrams, summary tables. The major regulatory requirements which must be respected (safety, hygiene, accessibility, performance) are based on the product and the nature of its implementation. Also, the proposed solutions to the risk that could threaten the safety or health of users are discussed and clearly illustrated with examples. To sum up, these guides provide an exhaustive description of the available knowledge helping to verify the conformity of a product to the regulations: definition of the risks, diagnosis, technical solutions, choices, implementation, compatibility with existing rules, controls, usage examples and useful links.

### 2.1 Classification of guides

The guides provided for our work contain information to verify the validity of technical documents provided by the manufacturers. They are made in a way to simplify the understanding and application of the law. Also, they include all the structural and dimensional characteristics related to the validation of the construction products.

However, our goal is to organize these guides on a query base that can be exploited by a system. The organization of this base will be according to classification criteria issued from the structure of the guides. We propose a classification according to three criteria:

- Organization by documentary source: we classify a query according to the type and number of the regulatory document from which it arises and the guides which reference it. Example: Guide "Coverage and Tile", outcome of DTU 40.211, 40.23, 40.22, 40.21.
- Organization by domain: Each request or set of queries is related to a specific application domain. Example: Accessibility, Security, etc.
- Organization by theme: This organization is the most important; we can classify queries according to technical solutions referenced in the guide. Example: Tile, Electricity, etc.

## 3 MODELLING OF REGULATORY CONSTRAINTS

### 3.1 Regulatory constraints of guides

Regulations that validate technical documents are first defined and represented in a language understandable by humans (experts and non-experts). In guides, the regulations have legal status derived from DTU. After reviewing various studies related to technical regulations, we have identified a structure to implement a process functionally complete to verify technical documents.

We use the guide "The coverage and tile" issued by CSTB in our work as a study model. Our first goal is to build a domain ontology based on the terms identified in the thread of the technical documents and contained in this guide. This ontology captures the different types of tile and their characteristics. Also, it defines their conditions of implementation and verification of various criteria such as: slope of tiles, support or climatology of where they should be installed. Monitoring of these instructions is drastic since the non-compliance with a requirement leads to a non-validation of the technical document. Each tile has a shape and a material according to these criteria. We were able to identify nine different types of tiles. In addition to its material and its form, each tile has an implementation that includes a slope support, an installation and an attachment. These characteristics are different from one tile to another hence the difference of their types:

- The slopes: the minimum allowable slopes for the types of tiles are given in the guides according to application areas and specific situations of exposition.
- Establishment of support: the tiles are based on a wooden litonnage which the elements are fixed at the rate of a point or a clip at each intersection of a chevron and a batten.
- Fixation: it is designed to maintain the assembly of the tiles with each other's when the wind effects risk to disturb the schedule. The fixing of tiles, if not total, is in a distributed manner, by nailing, and / or crocheting.

A convention must exist between the regulations (verification requests) and the structural properties of the Technical Documents. In our ontology, we considered this convention and integrated the content of the guides.

## 3.2 Construction of the technical document ontology OntoDT

Many solutions have been proposed for the manual construction of ontology among them (Gruninger and Fox, 1995), (Ushold and King, 1995) Methontology (Gomez-Perez et al. 2003), On-To-Knowledge (Staab et al. 2001). The common idea in these solutions is to reuse as much as possible existing ontologies related to the domain. This is the approach we adopted: we do not address the problem of NLP, we build an ontology from expert's interviews and by reusing existing ontological resources.

Specifically, in our work, with the help of photovoltaic experts, we began to identify what knowledge to represent along with the desired level of accuracy. From the technical documents, we extracted the terms corresponding to different elements or components of the products to certify. Then, we associated concepts to these terms. Our first result was an ontology where each concept considered relevant to the creation of technical documents is identified and modeled. We have confronted our ontology with the REEF thesaurus of the building domain developed by CSTB. The idea of reusing this Thesaurus is to have a controlled vocabulary and to link or integrate the technical advices to the REEF (Bus et al. 2009).

By doing so, we adopted an approach similar to that of (Hernandez et al. 2006): reusing the domain thesaurus that required heavy design efforts for the development of new resources with a higher formal level. This principle is interesting insofar as it avoids building a new ontology from scratch. The design of ontologies from thesauri has the advantage of considering all the terms identified by experts as being representative of the domain.

We adapted this approach to our case study by reusing part of the transformation process proposed by (Hernandez et al. 2006) and merging the resulting ontology with our ontology of technical document. However, the resulting ontology is light, with a minimum of semantics; it contains general concepts of the building industry and it lacks of specific terms of the Photovoltaic industry. To answer our problem we need specific concepts issued from technical documents.

Our approach reflects this in three steps: (I) Extracting hierarchical relationships explicit in the REEF thesaurus, (II) Removing redundancy in hierarchical relations of the thesaurus, (III) Merging the REEF ontology resulting of steps (I) and (II) with our ontology of technical documents.

### 3.2.1 Extracting concepts and relations

For this, we use the resources of the French REEF thesaurus, containing the major entities in the building industry. For example, the central concept of "étanchéité" (sealing) is connected to several more specific concepts (narrower) as "joint d'étanchéité" (sealing gasket) or "étanchéité à l'air" (airtightness). It is also tied by a relationship "broad" (broader term) to the concept of "calfeutrage" (caulking).

First, we hierarchically organize concepts corresponding to terms in the REEF thesaurus with the "subclassOf" relation. Some hierarchical relationships between concepts are directly acquired from explicit links present in the thesaurus: relations "narrower" and relations "broader" in the thesaurus. They are selected as candidates for eliciting a hierarchical relation between the concepts in the ontology corresponding to the related terms in the thesaurus.

If not formalized, redundancies in the structure of the thesaurus may exist. Our resulting ontology therefore initially contains some redundancy relative to the transitivity of the relations: if A is a subclass of B and B is a subclass of C, then A is a subclass of C: the hierarchical relationship between A and C need not be explicit. To remove this type of redundant relationships, we checked the relevance of each of the relations. Hernandez proposes an automatic analysis through graph theory operations.

### 3.2.2 Merging of the REEF ontology and the ontology of technical document

Once the REEF thesaurus has been transformed into the now so-called "REEF ontology" following the two steps described above, we integrate this ontology with the concepts of the ontology of the Technical document. We aim to explicit the semantics of the REEF terms describing the technical documents relative to the regulatory aspect of the photovoltaic field. The resulting ontology unifies and replaces the original ontology.

The most common approaches for merging ontologies use union or intersection to connect the resulting ontology to the original ontologies. In the union approach, the resulting ontology contains the union of entities coming from the original ontologies and suppose resolved the differences of representation of the same concept. In the intersection approach, the resulting ontology contains only the common parts of the original ontologies. We adopted the intersection approach: once the merge is completed, the resulting ontology contains:

- The intersection between the REEF ontology and the ontology of technical document
- Concepts specific to the Technical Document.

Several approaches and systems for merging ontologies have been proposed, including

PROMPT (Noy et al. 2000), Chimaera (Mcguinness et al. 2000), OntoMerge (Dou et al. 2002). We chose PROMPT as a tool for merging the REEF ontology and the technical document ontology as it creates a complete ontology. It identifies a set of operations for ontology merging (fusion of classes, merging of links) and a set of possible conflicts resulting from the application of these operations (name conflicts, redundancy in the class hierarchy).

```xml
<owl:Class rdf:about="#VerrePolymere">
  <rdfs:subClassOf rdf:resource="#ModulePhotoV"/>
  <rdfs:subClassOf>
    <owl:Class>
      <owl:intersectionOf rdf:parseType="Collection">
        <owl:Restriction>
          <owl:onProperty rdf:resource="#hasComponent"/>
          <!-- CableElectrique -->
          <owl:someValuesFrom rdf:resource="http://www.cstb.fr/reef/#01573"/>
        </owl:Restriction>
        <owl:Restriction>
          <owl:onProperty rdf:resource="#hasComponent"/>
          <!-- Cadre -->
          <owl:someValuesFrom rdf:resource="http://www.cstb.fr/reef/#01593"/>
        </owl:Restriction>
        …
      </owl:intersectionOf>
    </owl:Class>
  </rdfs:subClassOf>
</owl:Class>
```

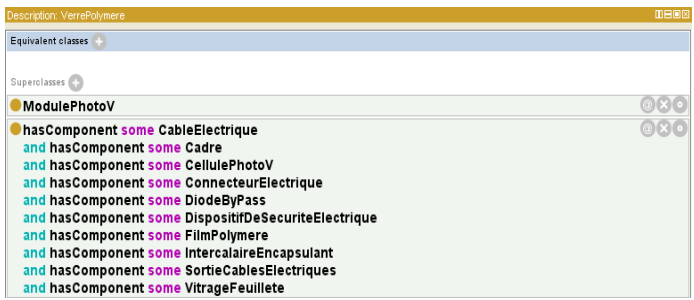

Figure 1: Example of a defined concept

As a result, our ontology of technical document called OntoDT has 121 classes and 39 properties formalized in the OWL Lite language. 35% of these classes are created from REEF terms. The remaining 65% are concepts more specific than those of the REEF thesaurus which only contains general concepts of the building industry. In its current state, it lacks specific terms relative to a particular field (e.g. Photovoltaic). However, it remains in constant evolution.

### 3.3 Integration of guides into OntoDT

We use the same approach to formalize information collected from the guide "Coverage and Tile". Each tile is represented as a concept and integrated into OntoDT. The ontology includes all the semantics that reflects the structural and dimensional criteria of a tile, these criteria are represented by properties.

To model each type of tile with its characteristics, we used the notion of axiom to represent this structure. This knowledge contains specific information to a particular area, they are called "domain axioms".

The modeling as an axiom allows us to describe general knowledge. It includes the definition of concepts and properties.

The following axiom provides a definition

$PlatClayTile \equiv Tile \wedge hasaMaterial.Clay \wedge hasAForm.Flat \wedge etc$

## 4 MODELLING GUIDES

We aim to show how it is possible to model the guides by representing the semantics of these rules. This knowledge is extracted manually from texts which they are derived. We will use for this SBVR formalism. It is based on a controlled vocabulary of the ontology OntoDT.

### 4.1 SBVR "Semantics of Business Vocabulary and Business Rules"

SBVR stands for "Semantics of Business Vocabulary and Business Rules". It is an OMG standard whose ultimate objective is to provide a meta-model that allows establishing data exchange interfaces for tools that create, organize, analyze and use vocabularies and business rules (Chapin et al. 2005). The SBVR meta-model facilitates the validation, analysis, alignment, and fusion of business rules for different tools of different constructors.

The development of an SBVR repository is done in two steps: the development of a business vocabulary and the writing of business rules based on the terms and concepts defined in the vocabulary. SBVR controlled vocabularies consist in hierarchies of concepts specific to some domains, their relationships, definitions and synonyms. In our case, the SBVR vocabulary is represented in Technical document ontology OntoDT.

However, technical standards can be understood in different ways that is why the manual intervention of a domain expert is essential. We argue that NLP approaches of knowledge extraction from regulatory texts can significantly alleviate the task of domain experts but cannot replace them. In our work, we do not consider linguistic analysis of texts and focus on the representation of expert knowledge. CSTB experts helped us to identify and classify the constraints expressed in the photovoltaic standards and then the rules which represent them. The goal of this categorization is to determine the levels of interoperability of each sentence or paragraph of the standards and classify it.

Once these texts are identified, a step of disambiguation is necessary. The transformation of texts into SBVR rules will provide a normative, unambiguous and reusable source.

The extraction of rules from standards or statutory text is a tedious job, it often requires to the structure the information. The regulations used have been detailed enough to show how their content can be converted into SBVR vocabulary and business rules. However, a clarification of the text was needed before the transformation into SBVR.

Example:

Let us consider the following example:

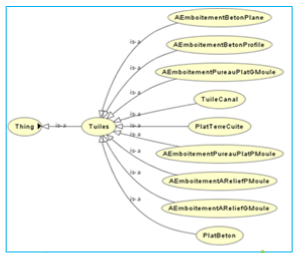

This table is extract from the guide "coverage and tile" expresses the acceptable slope for the plat till made with clay. We reformulate this table into set of SBVR rules

In our eg of SBVR rules we use several font styles:
- Concept are underlined in green color
- Properties are underlined in blue color
- Literal values are underlined in red
- Nouns and other information are given in orange

The concepts identified in this fragment are Tile, Zone1, Protected, which belong to the ontology of technical document.

### 4.2 Implementation of SBVR rules

In certain implementation process is based on the interpretation of the expert and its translation of knowledge into computer language. In other cases the logic of statements of human language is formally interpreted and later translated into language processable.

However, our goal is to model the way where experts use CSTB guides and try to automate their know-how. It forces us to follow their interpretation and to establish a formal representation of regulations (requests or rules). SBVR describes the concepts and requirements regardless of their implementations.

To validate our model, we developed a rule base to verify the compliance of DT with regulatory standards expressed in guides. The availability of verifications information maybe addressed as follows:
- Provide explicit information in the DT model. While this is natural for some aspects (material, form), others aspects are derived from more basic information and rely on fallible human judgment that are current causes of error (slope, fastening, fixing).
- Requests for verification are important they apply on properties that require more analysis like structural, dimensional or climatological constraints.

Our SBVR rules are implemented in SPARQL language. It allows reasoning automatically on the representation of regulatory and modelling the process of compliance check. This analysis model must implement a set of verification queries associated to a model of DT and to process of compliance check.

We organized the basic SPARQL queries representing different regulatory constraints in a process of compliance check. For each elementary component involved in a DT we associate a verification process consisting of SPARQL queries sequences that verify the compliance of its characteristics. This RDF representation called "elementary process" contains in addition to SPARQL "ASK" queries an SBVR rule and other elementary process (if the element contains several components in its nomenclature) Figure 2. For example: the elementary process of compliance testing slopes eligible for a tile contains SPARQL queries of compliance checking the slope and a SBVR representation of these regulatory constraints.

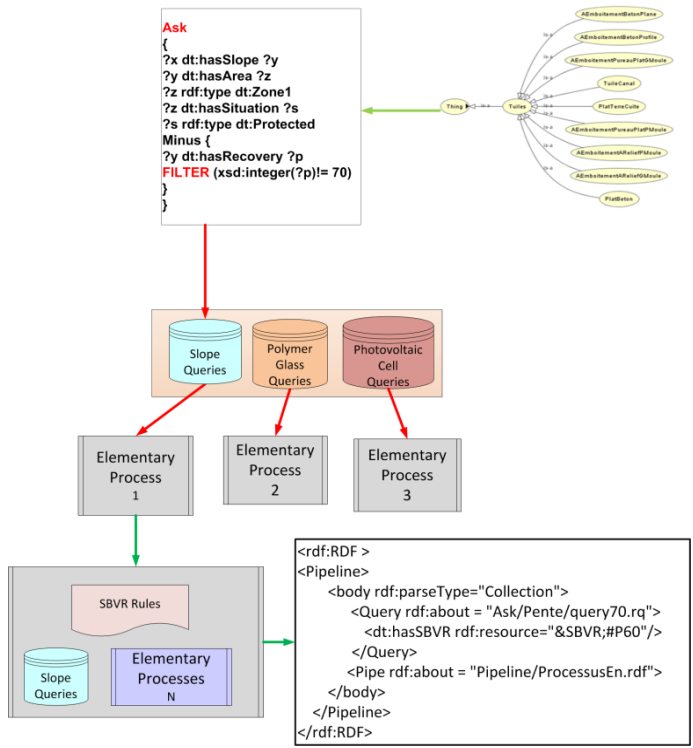

Figure 2: Extract of the process of compliance check

A complex process of verification is also defined recursively. It is built automatically from elementary processes associated with the components entering into its definition, under OntoDT ontology. Let's take the photovoltaic module "polymer glass" as an example, this module contain in its nomenclature: photovoltaic Cell, polymer glass and should be implemented on tiles.

The complex process of this module should contain elementary process of: photovoltaic Cell, polymer glass and tile.

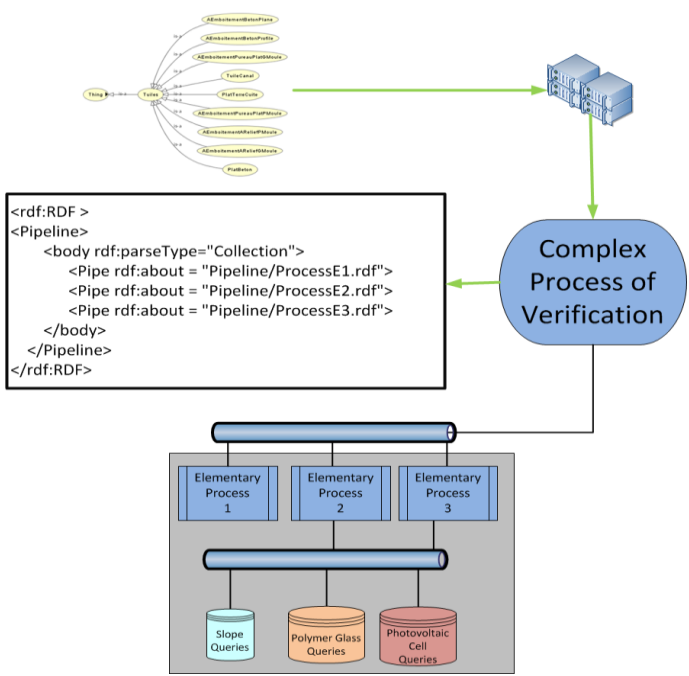

Figure 3: Example of the process of compliance check

This declarative representation as RDF annotation uses a resolution formalism, which allows monitoring, coordinating and describing an execution sequence in which we can run one or several queries sequentially with an abstract syntax. This controlled vocabulary is supported by the engine Kgram (Corby et al. 2010) and consists of four properties (body, if, then, else) and 7 classes where the class "Pipeline" is the main class that describes the beginning of annotation verification.

As shown below, this vocabulary can load, execute a set of rule or request and run on-demand queries/rules when a specific condition is met. Also, it can invoke other processes including the process itself. This notion of recursion allows to model complex process of solving where each process contains a description of one or several SPARQL queries.

- **Pipe:** Invocation of a new resolution process.
- **Load**: Load an annotation or ontology.
- **Query / Update:** Running a SPARQL query.
- **RuleBase:** Running a rule base.
- **Rule:** Running rules.
- **Test:** Refers to a condition (if-then-else), can run one/several rules/query or invoke them when a specific condition is met.

### 4.3 Supporting Technical Assessment

The implementation of our compliance checking model is based upon the matching of standards representations with those of a DT, i.e. the matching of SPARQL queries with RDF annotations if there are conditions for applying the standard. For this purpose, we use the KGRAM semantic engine.

One major problem when automating the compliance checking process is to justify the decisions taken by the system – the compliance or noncompliance of the product. Guides (Technical standards) that validate the products are modelled as set of compliance checking requests. The fail of these requests means the noncompliance of some components and involves a non-validation of the DT. The key point is to identify the source of this noncompliance by demonstrating the "Why" of this noncompliance and explain this negative decision to the industrials.

We use the KGRAM engine that implements the notions of "Event" and "Event Listener" allowing them to catch some predefined events in the inference engine. Some of these events are related to success or failure of execution of SPARQL queries.

In our case, we implement a program that's designed to intercept and identify which events are generated after the execution of the SPARQL queries. Thus we can identify the real cause of non-compliance by identifying the request failed and the cause of this failure. Once the condition is checked, non-compliance identified, a SPARQL query pattern is executed Figure 4 with as a parameter the ID of the non-compliant component. The results are one/several SBVR formulations related to regulations that validate this component. We sent these results as a noncompliance report to the users (industrials). In our case it is essentially the "why" of the negative decision.

```
SELECT DISTINCT ?z WHERE {
?x dt:hasSBVR ?y "
?b dt:hasSBVRrule ?z
FILTER (?y = ?b)
FILTER(?x = <"Rnon-conformité(idComposant)">
```

Figure 4: Extract of the request pattern

We use a knowledge base containing a list of justification established by experts in Figure 5. Each justification or answer is unique to a single state of non-compliance. The answer is extracted using another query pattern that takes as input the non-compliant component and displays the appropriate response.

```
<rdf:RDF>
<SBVR rdf:ID="P70">
<hasSBVRrule>
If a till is build in Zone 1 and has situation equal to protected and has recovery equal to 8 cm then it has slope equal to 70%
</hasSBVRrule>
</SBVR>
</rdf:RDF>
```

Figure 5: Extract of SBVR response

To summarize the process is triggered when a non-compliance of DT is identified. The verification of the parameters of non-compliance is achieved in Kgram engine. We identify the failing part of the query by retrieving the relative event. Each rule in this control process has a semantic link to a SBVR rules to explain the cause of non-compliance.

## 5 RELATED WORK

Regulatory modelling becomes an important issue. This issue was discussed under two different approaches. The first is to automatically analyse the regulations and to confront the complexity of natural language (Dinesh et al. 2008). In the second, regulatory constraints are written in normalized manner with the expert's help, which facilitates their translation into formal language (Reeder et al. 2007), (Nazarenko et al. 2011).

Our approach provides a third way, it takes account the regulations written in natural language, offers a tool for writing technical regulatory and automatically analyses the content of these documents.

In parallel, various efforts have been made to apply conformity rules on IFC representation or textual descriptions (Yurchyshyna et al. 2009), (Pauwels et al. 2011), (C. Eastman et al. 2009). The originality of our work is the combination of basic SPARQL queries and SBVR rules derived from regulation itself and use both to explain the why of decision making.

## 6 CONCLUSION

We are particularly interested in the modeling of the regulatory constraints derived from the Technical Guides and used to validate the Assessment. These Guides are regulatory complements offered by CSTB to the various industrials to enable easier reading of technical regulations. We formalize this GP in a machine-processable model to assist the creation of Technical Assessments by automating their validation.

First, we constructed a domain-ontology, which defines the main concepts involved in the Technical Guides. This ontology called "OntoDT" is coupled with domain thesauri. Several are being developed at CSTB among which one seems the most relevant by its volume and its semantic approach: the thesaurus from the REEF project.

Our second contribution is the use of standard SBVR (Semantics of Business Vocabulary and Business Rules) and SPARQL to reformulate the regulatory requirements of guides on the one hand in a controlled language and on the other hand in a formal language

Third, our model incorporates expert knowledge on the verification process of Technical Documents. We have organized the SPARQL queries representing regulatory constraints into several processes. Each component involved in the Technical Document corresponds to a elementary process of compliance checking. An elementary process contains a set of SPARQL queries to check the compliance of an elementary component. A full complex process for checking a Technical Document is defined recursively and automatically built as a set of elementary processes relative to the components which have their semantic definition in OntoDT.

Finally, we represent in RDF the association between the SBVR rules and SPARQL queries representing the same regulatory constraints. We use annotations to produce a compliance report in natural language to assist users in the writing of Technical Assessments.

As a result, we have designed a Semantic Web application to support and guide the process of writing Technical Assessment. The current version has allowed us to validate our approach. Also, we have developed a base of SBVR rules to describe business requirements of guides. This base of rules is implemented in SPARQL language. The experimentation and full evaluation, however, must be completed. In the future, we would like to enrich the OntoDT ontology and to compare our tool to a real use case